\documentclass[10pt,twocolumn,letterpaper]{article}

\usepackage{cvpr}
\usepackage{times}
\usepackage{epsfig}
\usepackage{graphicx}
\usepackage{amsmath}
\usepackage{amssymb}
\usepackage{enumitem} %lan
\usepackage{verbatim} % add by Zhizheng
\usepackage{xcolor} % add by lan
\usepackage{subfigure} % add by Zhizheng
\usepackage{multirow}
\usepackage{bigstrut}
\usepackage{booktabs}
\usepackage{comment}

\usepackage[pagebackref=true,breaklinks=true,letterpaper=true,colorlinks,bookmarks=false]{hyperref}

\cvprfinalcopy % *** Uncomment this line for the final submission

\def\cvprPaperID{2149} % *** Enter the CVPR Paper ID here

\ifcvprfinal\fi

\begin{document}

	%%%%%%%%% TITLE
	\title{Relation-Aware Global Attention for Person Re-identification}

	\author{Zhizheng Zhang{$^1$}\thanks{This work was done when Zhizheng Zhang was an intern at MSRA.} \qquad 
		Cuiling Lan$^2$\thanks{Corresponding author.}
		\qquad 
		Wenjun Zeng$^2$ 
		\qquad 
		Xin Jin$^1$
		\qquad
		Zhibo Chen$^{1\dagger}$
		\and 
		\normalsize
			$^1$University of Science and Technology of China \qquad $^2$Microsoft Research Asia\\
			{\tt\small \{zhizheng,jinxustc\}@mail.ustc.edu.cn} \quad {\tt\small \{culan,wezeng\}@microsoft.com} \quad {\tt\small chenzhibo@ustc.edu.cn}
			%		% For a paper whose authors are all at the same institution,
			%		% omit the following lines up until the closing ``}''.
			%		% Additional authors and addresses can be added with ``\and'',
			%		% just like the second author.
			%		% To save space, use either the email address or home page, not both
			%		\and
			%		\\
			%		Institution2\\
			%		First line of institution2 address\\
			%		{\tt\small secondauthor@i2.org}
			%		\and
	}

	\maketitle
	\thispagestyle{empty}

%%%%%%%%% ABSTRACT
\begin{abstract}
    For person re-identification (re-id), attention mechanisms have become attractive as they aim at strengthening discriminative features and suppressing irrelevant ones, which matches well the key of re-id, \it{i.e.}, discriminative feature learning. Previous approaches typically learn attention using local convolutions, ignoring the mining of knowledge from global structure patterns. Intuitively, the affinities among spatial positions/nodes in the feature map provide clustering-like information and are helpful for inferring semantics and thus attention, especially for person images where the feasible human poses are constrained. In this work, we propose an effective Relation-Aware Global Attention (RGA) module which captures the global structural information for better attention learning. Specifically, for each feature position, in order to compactly grasp the structural information of global scope and local appearance information, we propose to stack the relations, \it{i.e.}, its pairwise correlations/affinities with all the feature positions (\eg, in raster scan order), and the feature itself together to learn the attention with a shallow convolutional model. Extensive ablation studies demonstrate that our RGA can significantly enhance the feature representation power and help achieve the state-of-the-art performance on several popular benchmarks. The source code is available at {\small \url{ https://github.com/microsoft/Relation-Aware-Global-Attention-Networks}}.

\end{abstract}

%%%%%%%%% BODY TEXT
\section{Introduction}

Person re-identification (re-id) aims to match a specific person across different times, places, or cameras, which has drawn a surge of interests from both industry and academia. The challenge lies in how to extract discriminative features (for identifying the same person and distinguishing different persons) from person images where there are background clutter, diversity of poses, occlusion, \etc. 

\begin{figure}[ht]
	\begin{center}
		%\vspace{-3mm}
		\includegraphics[width=0.99\linewidth]{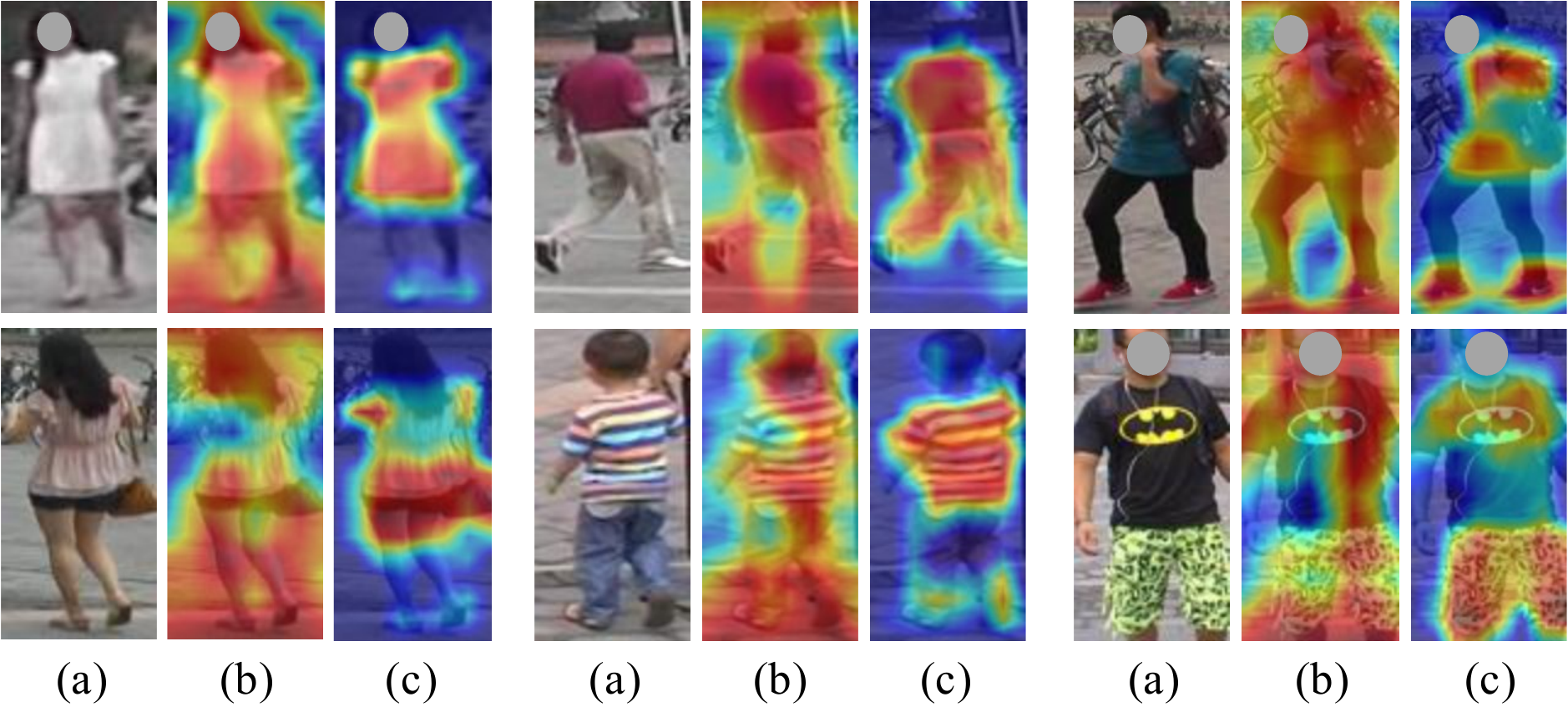}
	\end{center}
	\vspace{-2mm}
	\caption{Comparison of the learned attention between (b) spatial attention of CBAM \cite{woo2018cbam} without exploring relations, and (c) our proposed attention which captures global scope relations and mines from such structural information. (a) The original image \protect\footnotemark[1].}
	%\vspace{-3mm}
	\label{fig:comparion}
\end{figure}

Recently, many studies resort to attention design to address the aforementioned challenges in person re-id by strengthening the discriminative features and suppressing interference \cite{li2018harmonious, xu2018attention, li2018diversity, fu2019sta, fang2019bilinear, chen2019mixed}. Most of the attentions are learned by convolutions with limited receptive fields, which makes it hard to exploit the rich structural patterns in a global scope. One solution is to use large size filters in the convolution layer \cite{woo2018cbam}. The other solution is to stack deep layers \cite{wang2017residual} which increases the network size greatly. Besides, the studies in \cite{luo2016understanding} show that the effective receptive field of CNN only takes up a fraction of the full theoretical receptive field. These solutions cannot ensure the effective exploration of global scope information (\eg, global scope contents and corresponding positional geometry) for effective person re-id.

\footnotetext[1]{All faces in the images are masked for anonymization.}

Moreover, the non-local neural network is proposed in \cite{wang2018non} to allow the collection of global information by weighted summation of the features from all positions to the target position, where the connecting weight is calculated by the pairwise relation/affinity. Actually, for a target feature position, its pairwise relations with all the feature nodes/positions could contain valuable structural information of a global scope, \eg, clustering-like pattern (through \emph{pairwise affinities} and \emph{position information}). However, the non-local network overlooks the exploration of such rich global information. It only simply uses the learned relations/affinities as the weights to aggregate the features. Such a deterministic manner of using relations (\ie, weighted sum) has weak mining capability and lacks sufficient adaptability. Cao \etal observe that the learned connecting weights of non-local block are target position invariant \cite{cao2019gcnet}, which is not as adaptive as expected. We believe it is important to mine knowledge from the relations through a modeling function and leverage such valuable global scope structural information to infer attention. 
%affluent

In this paper, we propose an effective Relation-Aware Global Attention (RGA) module to efficiently learn discriminative features for person re-id. RGA explicitly explores the global scope relations for mining the structural information (clustering-like information). This is helpful for implicitly inferring semantics and thus attention.  Fig.~\ref{fig:comparion} shows our learned attention on the person re-id images. Thanks to the introduction and mining of global scope relations, our attention can focus on the discriminative human body regions. As illustrated in Fig.~\ref{fig:attention}~(c), for each feature node, \eg, a feature vector of a spatial position on a feature map, we model the pairwise relations of this node with respect to all the nodes and compactly stack the relations as a vector (which represents the global structural information) together with the feature of the node itself to infer the attention intensity via a small model. In this way, we take into account both the appearance feature and its global scope relations, to determine the feature importance from a global view. This mechanism is also consistent with the perception of human in finding discriminative features: making a global scope comparison to determine the importance.
%especially, the structural patterns of person images are easy to be captured, since the feasible human poses are constrained (not infinite).}
%\textcolor{purple}{For person images, the structural patterns can be mined from relations since the feasible human poses are limited.}

% In this paper, we propose an effective Relation-Aware Global Attention (RGA) module to efficiently learn discriminative features for person re-id. RGA explicitly explores the global scope relations for mining the spatial structural information (clustering-like information). This is helpful for implicitly inferring some kind of semantics and thus attention. Fig. \ref{fig:comparion} shows our learned attention on the person re-id images. Thanks to the introduction and mining of global scope relations, our attention can focus on the discriminative human body regions. As illustrated in Fig.~\ref{fig:attention}~(c), for each feature node, \ie, a feature vector of a spatial position on a feature map, we model the pairwise relations of this node with respect to all the nodes and compactly stack the relations as a vector (which represents the global structural information) together with the feature of the  node itself to infer the attention intensity via a small model. In this way, we take into account both the appearance feature and its corresponding global scope relations, to determine the feature importance from a global view. This mechanism is also consistent with the perception of human in finding discriminative features - making a global scope comparison to determine the importance. 

In summary, we have made two major contributions:
%\begin{itemize}[noitemsep,nolistsep]
\begin{itemize}[leftmargin=*,noitemsep,nolistsep]
	
\item We propose to globally learn the attention for each feature node by taking a global view of the relations among the features. With the global scope relations having valuable structural (clustering-like) information, we propose to mine semantics from relations for deriving attention through a learned function. Specifically, for a feature node, we build a compact representation by stacking its pairwise relations with respect to all feature nodes as a vector and mine patterns from it for attention learning.
    % \item We propose to globally learn the attention for each feature node by taking a global view of the relations among the features. With the global scope relations having valuable structural (clustering-like) information, we propose to mine semantics from relations for deriving attention through a learned modeling function. Specifically, for a feature node, we build a compact representation by stacking its pairwise relations with respect to all feature nodes as a vector and mine patterns from it for learning attention.
	
\item We design a relation-aware global attention (RGA) module which compactly represents the global scope relations and derives the attention based on them via two convolutional layers. We apply such design to spatial (RGA-S) and channel dimensions (RGA-C) and demonstrate its effectiveness for person re-id.
%We design a relation-aware global attention (RGA) module based on the relation-based compact representation and several convolutional layers. We apply such design to spatial (RGA-S) and channel dimensions (RGA-C) and demonstrate its effectiveness for person re-id.
	
\end{itemize}
We conduct extensive ablation studies to demonstrate the effectiveness of the proposed RGA in finding discriminative features and suppressing irrelevant ones for person re-id. Our scheme empowered by RGA modules achieves the state-of-the art performance on the benchmark datasets CUHK03~\cite{li2014deepreid}, Market1501~\cite{zheng2015scalable}, and MSMT17\cite{wei2018person}. 
%We will release our code upon acceptance of this paper.

%-------------------------------------------------------------------------
\section{Related Work}

\subsection{Attention and Person Re-id} 
Attention aims to focus on important features and suppress irrelevant features. This well matches the goal of handling aforementioned challenges in person re-id and is thus attractive. Many works learn the attention using convolutional operations with small receptive fields on feature maps \cite{wang2018mancs,zhao2017deeply,li2018harmonious,chen2019mixed}. However, intuitively, to have a good sense of whether a feature node is important or not, one should know the features of global scope which facilitates the comparisons needed for decision. 
% Attention aims to focus on important features and suppress irrelevant features. This well matches the goal of handling aforementioned challenges in person re-id and is thus attractive for re-id. Many works learn the attention using convolutional operations with small receptive fields on feature maps \cite{wang2018mancs,zhao2017deeply,li2018harmonious,chen2019mixed}. However, intuitively, to have a good sense of whether a feature node is important or not, one should know the features of global scope which facilitates the comparisons needed for decision.

In order to introduce more contextual information, Wang \etal and Yang \etal stack many convolutional layers in their encoder-decoder style attention module to have larger receptive fields \cite{wang2017residual,yang2019attention}. Woo \etal use a large filter size of 7$\times$7 over the spatial features in their Convolutional Block Attention Module (CBAM) to produce a spatial attention map  \cite{woo2018cbam}. In \cite{zhang2019nonlocal}, a non-local block \cite{wang2018non} is inserted before the encoder-decoder style attention module to enable attention learning based on globally refined features. Limited by the practical receptive fields, all these approaches are not efficient in capturing the large scope information to globally determine the spatial attention.

Some works explore the external clues of human semantics (pose or mask) as attention or to use them to guide the learning of attention \cite{xu2018attention,song2018mask,suh2018part,zhao2017spindle}. The explicit semantics which represent human structures is helpful for determining the attention. However, the external annotation or additional model for pose/mask estimation is usually required.

In this paper, we intend to explore the respective global scope relations for each feature node to learn attention. The structural information in the relation representation which includes both \emph{affinity} and \emph{location information} is helpful for learning semantics and infer attention. 

\subsection{Non-local/Global Information Exploration} 
Exploration of non-local/global information has been demonstrated to be very useful for image denoising \cite{buades2005non,BM3DTIP07,Lefki2017non}, texture synthesis \cite{efros1999texture}, super-resolution \cite{glasner2009super}, inpainting \cite{barnes2009patchmatch}, and even high level tasks such as image recognition, object segmentation \cite{wang2018non} and action localization \cite{chen2019relation}. Non-local block in \cite{wang2018non} aims at strengthening the features of the target position via aggregating information from all positions. For each target position/node, to obtain an aggregated feature, they compute a weighted summation of features of all positions (sources), with each weight obtained by computing the pairwise relation/affinity between the source feature node and target feature node. Then, the aggregated feature is added to the feature of the target position to form the output. Cao \etal visualized the target position specific connecting weights of source positions and surprisingly observed that the connecting weights are \emph{not} specific to the target positions \cite{cao2019gcnet}, \ie, the vector of connecting weights is invariant to the target positions where a connecting weight from a source position is actually only related to the feature of this source position. Simplified non-local block \cite{cao2019gcnet} exploits such target position invariant characteristic and determines each connecting weight by the source feature node only, which achieves very close performance as the original non-local. Note that the aggregated feature vector which is added to each target position is thus the same for different target positions and there is a lack of target position specific adaptation. 

Even though non-local block also learns pairwise relations (connecting weights), the global scope structural information is not well exploited. They just use them as weights to aggregate the features in a deterministic manner and have not mined the valuable information from the  relations. Different from non-local block, we aim to dig more useful information from a stacked relation representation and derive attention from it through a learned model. Our work is an exploration on how to make better use of relations and we hope it will inspire more works from the research community.
%to explore the relations in the future.

% Even though non-local block also learns pairwise relations (connecting weights), the global scope structural information is not well \textcolor{purple}{exploited}. They just use them as weights to aggregate the features in a deterministic manner and have not mined the valuable information from the  relations. Different from non-local block, we aim to dig more useful information from a stacked relation representation and derive attention from it through a learned model. Our work is also an exploration on how to make better use of relations and we hope it will inspire more works which explore the relations in the future from the research community.  

%-------------------------------------------------------------------------
\section{Relation-Aware Global Attention}

For discriminative feature extraction in person re-id, we propose a Relation-aware Global Attention (RGA) module which makes use of the compact global scope structural relation information to infer the attention. In this section, we first give the problem formulation and introduce our main idea in Subsec.~3.1. For CNN, we elaborate on the designed spatial relation-aware global attention (RGA-S) in Subsec.~3.2 and channel relation-aware global attention (RGA-C) in Subsec.~3.3, respectively. We analyze and discuss the differences between our attention and some related approaches in Subsec.~3.4.

\begin{figure}[t]
	\begin{center}
		%\vspace{-3mm}
		\includegraphics[width=1\linewidth]{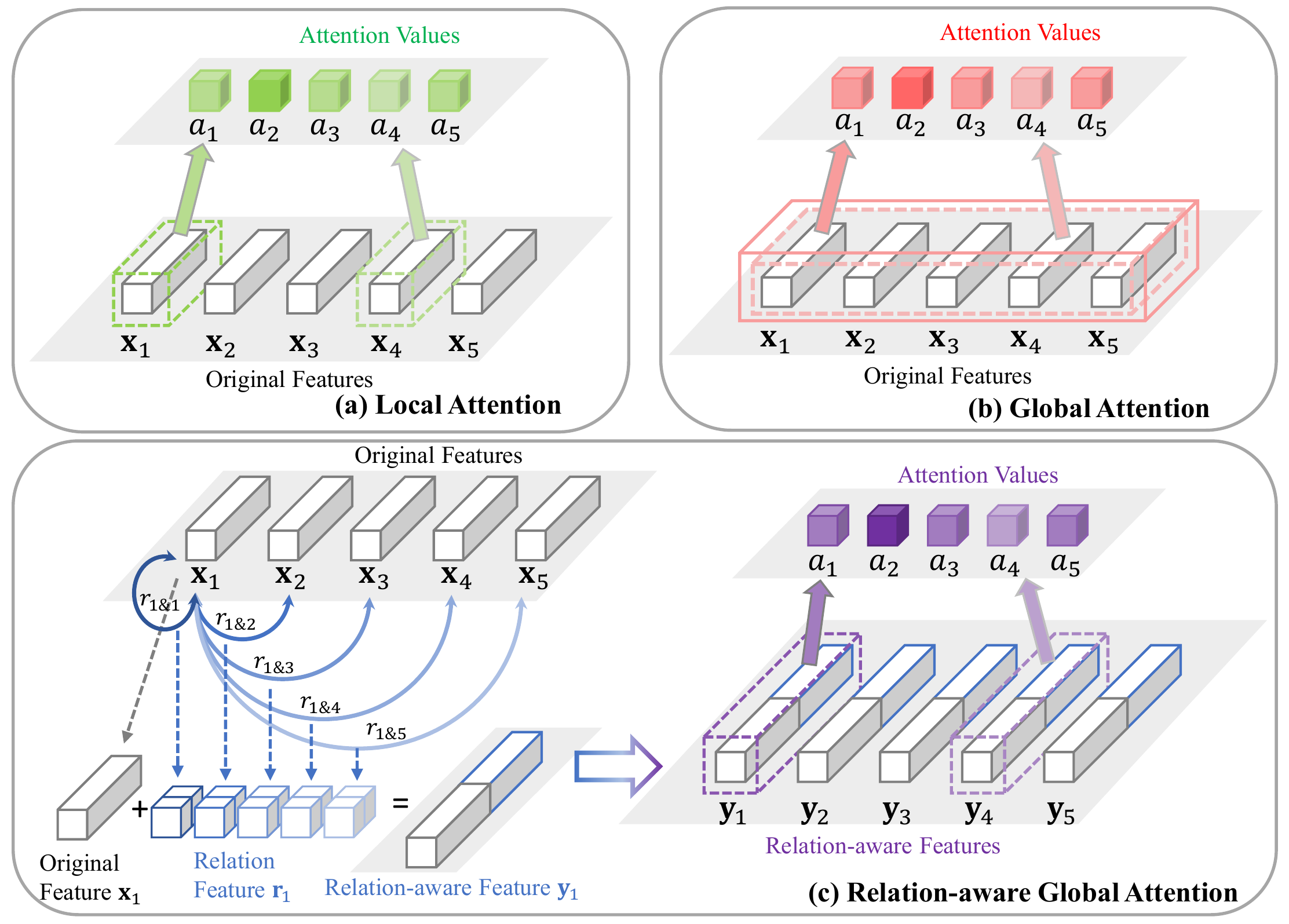}
	\end{center}
	\vspace{-1mm}
	\caption{Illustration of learning attention values $a_1,\cdots,a_5$ for five feature vectors/nodes $\mathbf{x}_1, \cdots,\mathbf{x}_5$. (a) Local attention: learn attention locally (\eg, based on individual feature as shown). (b) Global attention: learn attention jointly from all the 5 feature vectors (\eg, by concatenating them together). (c) Proposed relation-aware global attention: learn attention by taking into account the global relation information. For the $i^{th}$ (here $i=1$) feature vector, the global scope relation information is represented by stacking the pairwise relations $\mathbf{r}_i= [r_{i,1}, \cdots, r_{i,5}, r_{1,i}, \cdots, r_{5,i}]$. Note that $r_{i\&j}=[r_{i,j},r_{j,i}]$. Unlike (a) that lacks global awareness and (b) that lacks explicit relation exploration, our proposed attention is determined through a learned function with the global scope relations which contain structural information as input.}
	%which concatenates the features together
	\vspace{-3mm}
	\label{fig:attention}
\end{figure}

\begin{figure*}[t]
		\subfigure[Spatial Relation-Aware Global Attention] {\includegraphics[width=1\columnwidth]{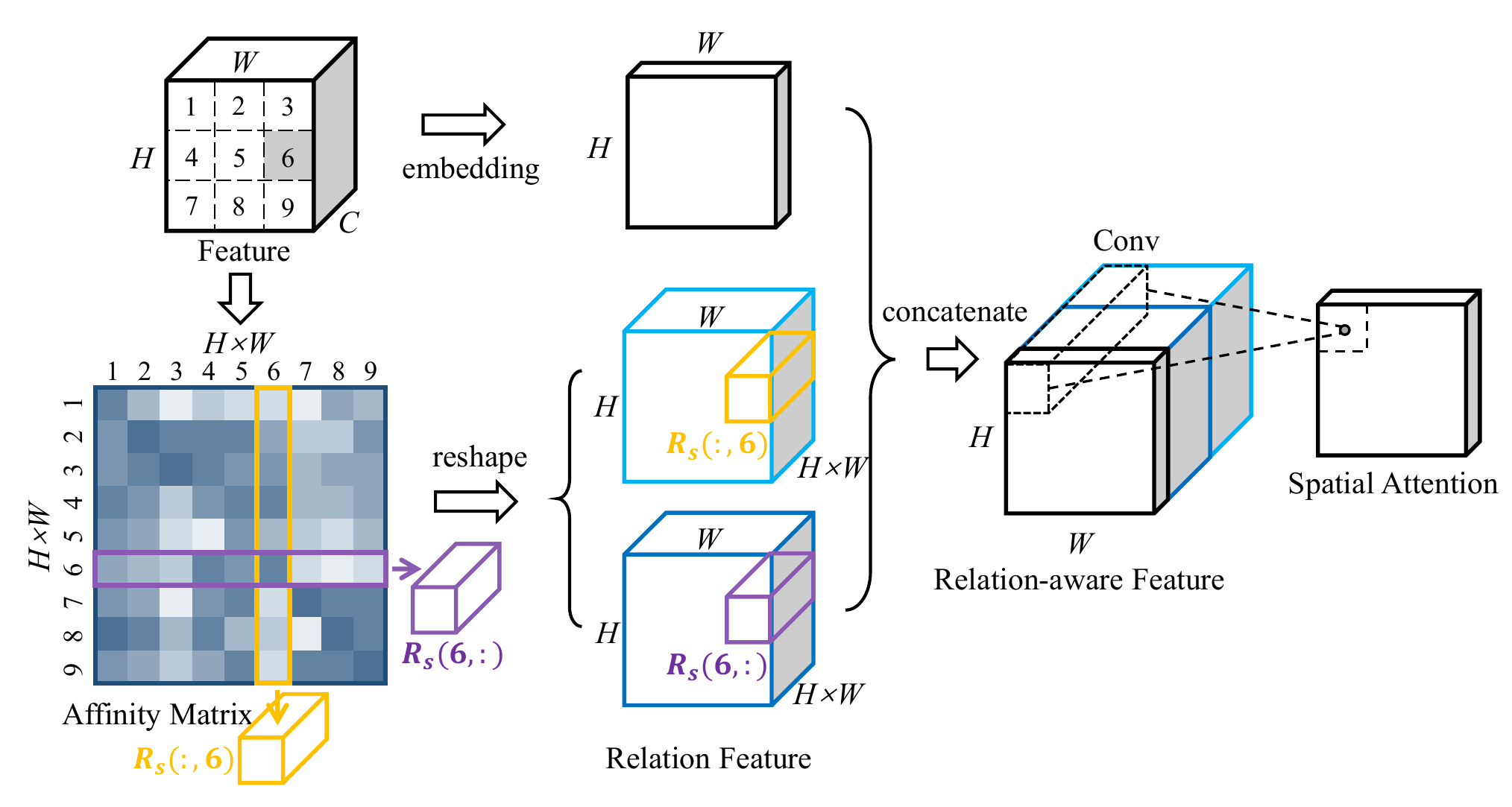}}
		\hspace{.3cm}
		\subfigure[Channel Relation-Aware Global Attention] {\includegraphics[width=1\columnwidth]{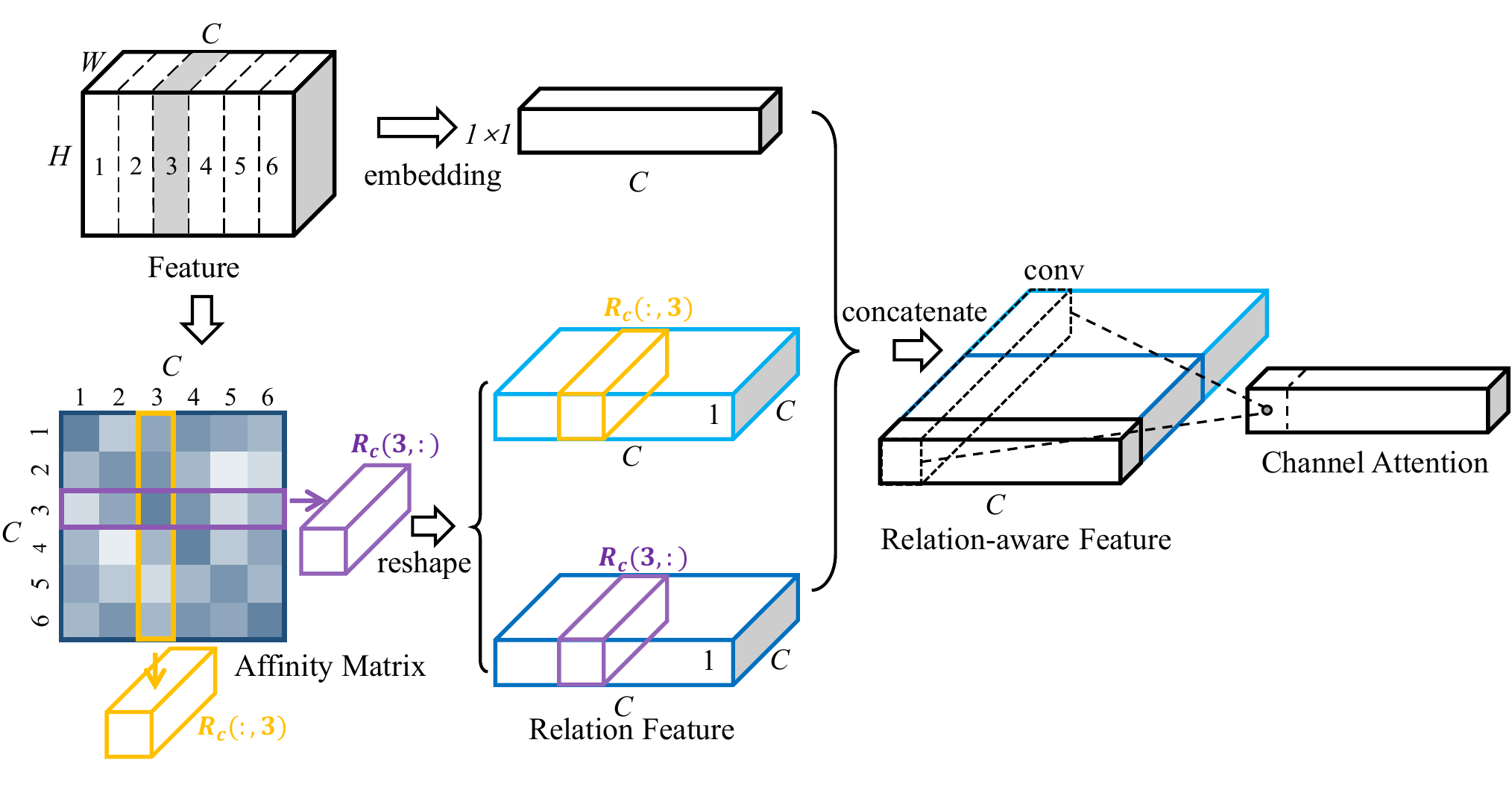}}
		\caption{Diagram of our proposed Spatial Relation-aware Global Attention (RGA-S) and Channel Relation-aware Global Attention (RGA-C). When computing the attention at a feature position, in order to grasp information of global scope, we stack the pairwise relation items, {\it{i.e.}}, its correlations/affinities with all the feature positions, and the unary item, {\it{i.e.}}, the feature of this position, for learning the attention with convolutional operations.}
		\label{fig:attention-cnn}
\end{figure*} 

\subsection{Formulation and Main Idea}
\label{subsec:idea}

Generally, for a feature set $\mathcal{V} = \{ \mathbf{x}_i \in \mathbb{R}^d, i = 1,\cdots,N\}$ of $N$ correlated features with each of $d$ dimensions, the goal of attention is to learn a mask denoted by $\mathbf{a} = (a_1,\cdots,a_N) \in \mathbb{R}^N$ for the $N$ features to weight/mask them according to their relative importance. Note that we also refer to a feature vector as feature node or feature.

Two common strategies are used to learn the attention value $a_i$ of the $i^{th}$ feature vector, as illustrated in Fig.~\ref{fig:attention} (a) and (b). \textbf{(a) Local attention}: the attention for a feature node is determined locally, \eg, applying a shared transformation function $\mathcal{F}$ on itself, \ie, $a_i = \mathcal{F}(\mathbf{x}_i)$ \cite{wang2018mancs}. However, such local strategies do not fully exploit the correlations from a global view and ignore the global scope structural information. For vision tasks, deep layers \cite{wang2017residual} or large-size kernels \cite{woo2018cbam} are used to remedy this problem. \textbf{(b) Global attention}: one  solution is to use all the feature nodes (\eg by concatenation) together to jointly learn attention, \eg, using fully connected operations. However, this is usually computationally inefficient and difficult to optimize, as it requires a large number of parameters especially when the number of features $N$ is large \cite{liu2018spatial}.

In contrast to these strategies, we propose a relation-aware global attention that enables i) the exploitation of global structural information and knowledge mining, and ii) the use of shared transformation function for different individual feature positions to derive the attention. For re-id, the latter makes it possible to globally compute the attention by using local convolutional operations. Fig.~\ref{fig:attention}~(c) illustrates our basic idea for the \textbf{proposed relation-aware global attention}. 
The main idea is to exploit the pairwise relation (\eg affinity/similarity) of the current ($i^{th}$) feature node with all the feature nodes, respectively, and stack them (with some fixed order) to compactly represent the global structural information for the current feature node. Specifically, we use $r_{i,j}$ to represent the affinity between the $i^{th}$ feature and the $j^{th}$ feature. For the feature node $\mathbf{x}_i$, its affinity vector is $\mathbf{r}_i = [r_{i,1},r_{i,2},\cdots,r_{i,N},r_{1,i},r_{2,i},\cdots,r_{N,i}]$. Then, we use the feature itself and the pairwise relations, \ie, $\mathbf{y}_i = [\mathbf{x}_i, \mathbf{r}_i]$, as the feature used to infer its attention through a \emph{learned} transformation function. Note that $\mathbf{y}_i$ contains global information.

Mathematically, we denote the set of features and their relations by a graph $G=\left(\mathcal{V}, \mathcal{E} \right)$, which comprises the node set $\mathcal{V}$ of $N$ features, together with an edge set $\mathcal{E}= \{ r_{i,j} \in \mathbb{R}, i = 1,\cdots,N $ and $j = 1,\cdots, N\}$. The edge $r_{i,j}$ represents the relation between the $i^{th}$ node and the $j^{th}$ node. The pairwise relations for all the nodes can be represented by an affinity matrix $R \in \mathbb{R}^{N \times N}$, where the relation between node $i$ and $j$ is $r_{i,j}$ = $R(i,j)$. $\mathbf{r}_i = [R(i,:),R(:,i)]$, where $R(i,:)$ denotes the $i^{th}$ row of $R$ and $R(:,i)$ denotes the $i^{th}$ column of $R$.

\textbf{Discussion}: For the $i^{th}$ feature node $\mathbf{x}_i$, its corresponding relation vector $\mathbf{r}_i$ provides a compact representation to capture the global structural information, \ie, both the position information and pairwise affinities with respect to all feature nodes. With the pairwise relation values denoting the similarity/affinity between every feature node and the current feature node while their locations in the relation vector denoting the positions (indexes) of the feature nodes, the relation vector reflects \emph{the clustering states and patterns} of all the nodes with respect to the current node, which benefits the global determination of the relative importance (attention) of $\mathbf{x}_i$. With such affluent structural information/patterns contained, we propose to mine from the relations for effectively learning attention through a modeling function. The structural patterns of person re-id images span in a learnable space considering the feasible poses are constrained by the human physical structure.

\subsection{Spatial Relation-Aware Global Attention}
\label{subsec:RGAS}

Given an intermediate feature tensor $X\in \mathbb{R}^{C\times H \times W}$ of width $W$, height $H$, and $C$ channels from a CNN layer, we design a spatial relation-aware attention block, namely RGA-S, for learning a spatial attention map of size $H \times W$. We take the $C$-dimensional feature vector at each spatial position as a feature node. All the spatial positions form a graph $G_s$ of $N=W\times H$ nodes. As illustrated in Fig.~\ref{fig:attention-cnn}~(a), we raster scan the spatial positions and assign their identification number as 1,$\cdots$, $N$. We represent the $N$ feature nodes as $\mathbf{x}_i\in \mathbb{R}^C$, where $i=1,\cdots,N$. 

The pairwise relation (\ie affinity) $r_{i,j}$ from node $i$ to node $j$ can be defined as a dot-product affinity in the embedding spaces as:
\begin{equation}\label{eq:rij}
r_{i,j}=f_s(\mathbf{x}_i, \mathbf{x}_j)=\theta_s(\mathbf{x}_i)^T\phi_s(\mathbf{x}_j),
\end{equation}	
where $\theta_s$ and $\phi_s$ are two embedding functions implemented by a $1\times 1$ spatial convolutional layer followed by batch normalization (BN) and ReLU activation, \ie $\theta_s(\mathbf{x}_i) = ReLU(W_{\theta}\mathbf{x}_i)$, $\phi_s(\mathbf{x}_i) = ReLU(W_{\phi}\mathbf{x}_i)$, where $W_{\theta}\in \mathbb{R}^{\frac{C}{s_1} \times C}$ and $W_{\phi}\in \mathbb{R}^{\frac{C}{s_1} \times C}$. $s_1$ is a pre-defined positive integer which controls the dimension reduction ratio. Note that BN operations are all omitted to simplify the notation. Similarly, we can get the affinity from node $j$ to node $i$ as $r_{j,i}=f_s(\mathbf{x}_j, \mathbf{x}_i)$. We use the pair $(r_{i,j}, r_{j,i})$ to describe the bi-directional relations between $\mathbf{x}_i$ and $\mathbf{x}_j$. Then, we represent the pairwise relations among all the nodes by an affinity matrix $R_s \in \mathbb{R}^{N\times N}$.

For the $i^{th}$ feature node, we stack its pairwise relations with all the nodes in a certain fixed order (\eg, raster scan order), \ie, node identities as $j\!=\!1,2,\cdots,N$, to obtain a relation vector $\mathbf{r}_i=[R_{s}(i,:),R_{s}(:,i)] \in \mathbb{R}^{2N}$. For example, as in Fig.~\ref{fig:attention-cnn}~(a), the sixth row and the sixth column of the affinity matrix $R_s$, \ie $\mathbf{r}_6 = [R_{s}(6,:)$, $R_{s}(:,6)]$, is taken as the relation features for deriving the attention of the sixth spatial position.

To learn the attention of the $i^{th}$ feature node, besides the pairwise relation items $\mathbf{r}_i$, we also include the feature itself $\mathbf{x}_i$ to exploit both the global scope structural information relative to this feature and the local original information. Considering these two kinds of information are not in the same feature domain, we embed them respectively and concatenate them to get the spatial relation-aware feature $\widetilde{\mathbf{y}_i}$:
\begin{equation}\label{eq:yi}
\widetilde{\mathbf{y}_i}=[pool_c(\psi_s(\mathbf{x}_i)), \varphi_s(\mathbf{r}_i)],
\end{equation}
where $\psi_s$ and $\varphi_s$ denote the embedding functions for the feature itself and the global relations, respectively. They are both implemented by a spatial $1\times1$ convolutional layer followed by BN and ReLU activation, \ie, $\psi_s(\mathbf{x}_i) = ReLU(W_{\psi}\mathbf{x}_i)$, $\varphi_s(\mathbf{r}_i) = ReLU(W_{\varphi}\mathbf{r}_i)$, where $W_{\psi}\in \mathbb{R}^{\frac{C}{s_1} \times C}$, $W_{\varphi}\in \mathbb{R}^{\frac{2N}{2s_1} \times 2N}$. $pool_c(\cdot)$ denotes global average pooling operation along the channel dimension to further reduce the dimension to 1. Then $\widetilde{\mathbf{y}_i} \in \mathbb{R}^{1+N/s_1}$. Note that other convolution kernel size (\eg 3$\times$3) can also be used. We found they achieve very similar performance so that we use $1\times1$ convolutional layer for lower complexity.

The global scope relations contain affluent structural information (\eg, clustering-like state in  feature space with semantics), we propose to mine valueable knowledge from them for inferring attention through a learnable model. We obtain the spatial attention value $a_i$ for the $i^{th}$ feature/node through a modeling function as:
\begin{equation}\label{eq:ai}
a_i=Sigmoid(W_2 ReLU (W_1\widetilde{\mathbf{y}_i})),
\end{equation}
where $W_1$ and $W_2$ are implemented by $1\times1$ convolution followed by BN. $W_1$ shrinks the channel dimension with a ratio of $s_2$ and $W_2$ transforms the channel dimension to 1.

\subsection{Channel Relation-Aware Global Attention}
\label{subsec:RGAC}

Given an intermediate feature tensor $X\in \mathbb{R}^{C\times H \times W}$, we design a relation-aware channel attention block, namely RGA-C, for learning a channel attention vector of $C$ dimensions. We take the $d = H\times W$-dimensional feature map at each channel as a feature node. All the channels form a graph $G_c$ of $C$ nodes. We represent the $C$ feature node as $\mathbf{x}_i\in \mathbb{R}^d$, where $i=1,\cdots,C$.

Similar to spatial relation, the pairwise relation $r_{i,j}$ from node $i$ to node $j$ can be defined as a dot-product affinity in the embedding spaces as:
\begin{equation}\label{eq:rijc}
r_{i,j}=f_c(\mathbf{x}_i, \mathbf{x}_j)=\theta_c(\mathbf{x}_i)^T\phi_c(\mathbf{x}_j),
\end{equation}	
where $\theta_c$ and $\phi_c$ are two embedding functions that are shared among feature nodes. We achieve the embedding by first spatially flattening the input tensor $X$ into $X'\in \mathbb{R}^{(HW)\times C\times 1}$ and then using a $1\times 1$ convolution layer with BN followed by ReLU activation to perform a transformation on $X'$. As illustrated in Fig.~\ref{fig:attention-cnn}~(b), we obtain and then represent the pairwise relations for all the nodes by an affinity matrix $R_c \in \mathbb{R}^{C\times C}$.

For the $i^{th}$ feature node, we stack its corresponding pairwise relations with all the nodes to have a relation vector $\mathbf{r}_i=[R_{c}(i,:),R_{c}(:,i)] \in \mathbb{R}^{2C}$, to represent the global structural information.
%\textcolor{green}{For example, as illustrated in Fig.~\ref{fig:attention-cnn}~(b), the third row and the third column of the affinity matrix $R_c$, \ie $\mathbf{r}_3=[R_{c}(3,:)$, $R_{c}(:,3)]$, is taken as the relation features for deriving the attention of the third channel node.}

To infer the attention of the $i^{th}$ feature node, similar to the derivation of the spatial attention, besides the pairwise relation items $\mathbf{r}_i$, we also include the feature itself $\mathbf{x}_i$. Similar to Eq.~(\ref{eq:yi}) and (\ref{eq:ai}), we obtain the channel relation-aware feature $\mathbf{y}_i$ and then the channel attention value $a_i$ for the $i^{th}$ channel. Note that all the transformation functions are shared by nodes/channels. There is no fully connected operation across channels.

\subsection{Analysis and Discussion}
\label{subsec:discussion}
We analyze and discuss the differences from other related approaches. Moreover, we discuss the joint use of the spatial and channel RGA and their integration strategies.

\noindent \textbf{RGA vs. CBAM\cite{woo2018cbam}.} Most of the attention mechanisms in CNN are actually local attention, which determines the attention of a feature position using local context \cite{woo2018cbam,wang2017residual,wang2018mancs,li2018harmonious}. Taking the representative attention module CBAM \cite{woo2018cbam} as an example, it uses a convolution operation of filter size $7\times7$ followed by sigmoid activation function to determine the attention of a spatial feature position. Therefore, only $7\times 7=49$ neighboring feature nodes are exploited to determine the attention of the center position. In contrast, for our spatial RGA (RGA-S), for a spatial feature position, we jointly exploit the feature nodes at all spatial positions to globally determine the attention. We achieve this through simple $1\times 1$ convolutional operations on the vector of stacked relations.  

\noindent \textbf{RGA vs. Non-local~(NL) \cite{wang2018non} and Simplified NL \cite{cao2019gcnet}.} Non-local block \cite{wang2018non} exploits the global context to refine the feature at each spatial position. For a target feature position, to obtain an aggregated feature which is then added to the original feature for refinement, they compute a weighted summation of features of source positions. Even though there is structural information from the pairwise relations, non-local overlooks the exploration of such valuable information and only uses the relations as weights for feature aggregation through such a deterministic manner. As observed and analyzed by Cao \etal \cite{cao2019gcnet}, the connecting weights in non-local block are \emph{invariant to the target positions}, with each connecting weight \emph{locally} determined by the source feature node itself. Therefore, the vector of the connecting weights is the same for different target positions, so is the corresponding aggregated feature vector. This results in a lack of target position specific adaptation. In contrast, in our RGA, even though we similarly make use of the pairwise relations, our intention is rather different which is to \emph{mine} knowledge from the global scope structural information of the relations through a learned modeling function. 
%representations

\noindent\textbf{Usage of RGA-S and RGA-C.} RGA-S and RGA-C can be plugged into any CNN network in a plug-and-play fashion. We can use RGA-S or RGA-C alone, or jointly use them in sequence (\eg, apply RGA-C following RGA-S which is denoted as RGA-SC) or in parallel (RGA-S//C).

%-------------------------------------------------------------------------
\section{Experiments}
\label{subsec:results-reid}

\subsection{Implementation Details and Datasets}
\noindent\textbf{Network Settings}. Following the common practices in re-id \cite{zhang2017alignedreid,almazan2018re,zhang2019densely}, we take ResNet-50 \cite{he2016deep} to build our baseline network and integrate our RGA modules into the RetNet-50 backbone for effectiveness validation. Similar to \cite{sun2017beyond,zhang2019densely}, the last spatial down-sampling operation in the conv5\_x block is removed. In our experiments, we add the proposed RGA modules after all of the four residual blocks (including conv2\_x, conv3\_x, conv4\_x and conv5\_x). For brevity, we also refer to the scheme as \emph{RGA}. Within RGA modules, we set the ratio parameters $s_1$ and $s_2$ to be 8. We use both identification (classification) loss with label smoothing \cite{szegedy2016rethinking} and triplet loss with hard mining \cite{hermans2017defense} as supervision. Note that we do not implement re-ranking \cite{zhong2017re}.
%for clear comparisons in all our experiments. 
%More details please see our Supplementary.

\noindent\textbf{Training}. We use the commonly used data augmentation strategies of random cropping \cite{wang2018resource}, horizontal flipping, and random erasing \cite{zhong2017random,wang2018resource,wang2018mancs}. The input image size is $256\times128$ for all the datasets. The backbone network is pre-trained on ImageNet~\cite{deng2009imagenet}. We adopt the Adam optimizer to train all models for 600 epochs with the learning rate of $8\times 10^{-4}$ and the weight decay of $5\times 10^{-4}$. %Please refer to Supplementary for more details.

\noindent\textbf{Datasets and Evaluation Metrics}.
We conduct experiments on three public person re-id datasets, \ie, CUHK03 \cite{li2014deepreid}, Market1501 \cite{zheng2015scalable}, and the large-scale MSMT17 \cite{wei2018person}. We follow the common practices and use the cumulative matching characteristics (CMC) at Rank-1 (R1) and mean average precision (mAP) to evaluate the performance.
%To be consistent with what were done in prior works for performance comparisons,

% \textcolor{purple}{To be consistent with what was done in prior works for performance comparisons,} we conduct experiments on four public person re-id datasets, \ie, CUHK03 \cite{li2014deepreid}, Market1501 \cite{zheng2015scalable}, DukeMTMC-reID \cite{zheng2017unlabeled} and the large-scale MSMT17 \cite{wei2018person}. For CUHK03, we use the new protocol as in \cite{zhong2017re,zheng2018pedestrian,he2018recognizing}. We follow the common practices and use the cumulative matching characteristics (CMC) at Rank-1 (R1) and mean average precision (mAP) to evaluate the performance.
%More details about datasets can be found in our supplementary. 

\subsection{Ablation Study}
Following the common practice, we perform the ablation studies on two representative datasets CUHK03 (with the Labeled bounding box setting) and Market1501.

\noindent\textbf{RGA related Models vs. Baseline.}
Table~\ref{tab:Relation} shows the comparisons of our spatial RGA (\emph{RGA-S}), channel RGA (\emph{RGA-C}), their combinations, and the baseline. We observe that:

\noindent\textbf{1)} Either \emph{RGA-S} or  \emph{RGA-C} significantly improves the performance over \emph{Baseline}. On CUHK03, \emph{RGA-S}, \emph{RGA-C}, and the sequentially combined version \emph{RGA-SC} significantly outperform \emph{Baseline} by \textbf{5.7\%}, \textbf{6.6\%}, and \textbf{8.4\%} respectively on mAP, and \textbf{5.5\%}, \textbf{5.5\%}, and \textbf{7.3\%} respectively on Rank-1 accuracy. On Market1501, even though the performance of \emph{Baseline} is already very high, \emph{RGA-S} and \emph{RGA-C} improve the mAP by 3.8\% and 4.2\%, respectively.

\noindent\textbf{2)} For learning attention, even without taking the visual features (Ori.), \ie, feature itself, as part of the input, using the proposed global relation representation itself (\emph{RGA-S w/o Ori.} or \emph{RGA-C w/o Ori.}) significantly outperforms \emph{Baseline}, by \eg 5.0\% or 5.9\% in mAP accuracy on CUHK03.

\noindent\textbf{3)} For learning attention, without taking the proposed global relation (Rel.) as part of the input, the scheme \emph{RGA-S w/o Rel.} or \emph{RGA-C w/o Rel.} are inferior to our scheme \emph{RGA-S} or \emph{RGA-C} by 2.4\% or 1.9\% in mAP accuracy on CUKH03. Both 2) and 3) demonstrate that global scope relation representation is very powerful for learning attention.

\noindent\textbf{4)} The combination of the spatial RGA and channel RGA achieves the best performance. We study three ways of combination: parallel with a fusion (\emph{RGA-S\///C}), sequential spatial-channel (\emph{RGA-SC}), sequential channel-spatial (\emph{RGA-CS}). \emph{RGA-SC} achieves the best performance, 2.7\% and 1.8\% higher than \emph{RGA-S} and \emph{RGA-C}, respectively, in mAP accuracy on CUHK03. Sequential architecture allows the later module to learn attention based on modulated features resulting from its preceding attention module, which makes the optimization easier.

\begin{table}[t]%htbp]
	\centering
	\tabcolsep=6.5pt
	\newcommand{\tabincell}[2]{\begin{tabular}{@{}#1@{}}#2\end{tabular}}
	\caption{Performance (\%) comparisons of our models with the baseline, and the effectiveness of the global relation representation (Rel.) and the feature itself (Ori.). w/o: without.}
	\footnotesize
	% 		\scriptsize
	\resizebox{0.48\textwidth}{!}{
		\begin{tabular}{cccccc}
			\hline
			\multicolumn{2}{c}{\multirow{2}[4]{*}{Model}} & \multicolumn{2}{c}{CUHK03(L)} & \multicolumn{2}{c}{Market1501} \bigstrut\\
			\cline{3-6}    \multicolumn{2}{c}{} & R1 & mAP & R1 & mAP \bigstrut\\
			\hline Baseline
			& \multicolumn{1}{l}{ResNet-50} & 73.8 & 69.0 & 94.2 & 83.7 \bigstrut\\
			\hline
			\multirow{3}[2]{*}{Spatial} & \multicolumn{1}{l}{RGA-S w/o Rel.} & 76.8 & 72.3 & 94.3  & 83.8 \bigstrut[t]\\
			& \multicolumn{1}{l}{RGA-S w/o Ori.} &  78.2  &  74.0  &  95.4  & 86.7 \\
			& \multicolumn{1}{l}{RGA-S} & \textbf{79.3} & \textbf{74.7} & \textbf{96.0} & \textbf{87.5} \bigstrut[b]\\
			\hline
			\multirow{3}[2]{*}{Channel} & \multicolumn{1}{l}{RGA-C w/o Rel.} & 77.8 & 73.7 &  94.7  & 84.8 \bigstrut[t]\\
			& \multicolumn{1}{l}{RGA-C w/o Ori.} &  78.1  &  74.9  &  95.4  & 87.1 \\
			& \multicolumn{1}{l}{RGA-C} & \textbf{79.3} & \textbf{75.6} & \textbf{95.9} & \textbf{87.9} \bigstrut[b]\\
			\hline
			\multirow{3}[2]{*}{Both} & \multicolumn{1}{l}{RGA-S//C} & 77.3 & 73.4 & 95.3 & 86.6 \bigstrut[t]\\
			& \multicolumn{1}{l}{RGA-CS} & 78.6 & 75.5 & 95.3 & 87.8 \\
			& \multicolumn{1}{l}{RGA-SC} & \textbf{81.1} & \textbf{77.4} & \textbf{96.1} & \textbf{88.4} \bigstrut[b]\\
			\hline
		\end{tabular}%
	}
	\label{tab:Relation}%
	\vspace{-3mm}
\end{table}%

\noindent\textbf{RGA vs. Other Approaches}. For fairness of comparison, we re-implement their designs on top of our baseline and show the results in Table \ref{tab:OtherAttention}.

\noindent\textbf{1) Spatial attention.}  \emph{CBAM-S} \cite{woo2018cbam} uses a large filter size of 7$\times$7 to learn attention while \emph{FC-C} \cite{liu2018spatial} uses fully connection over the (channel-pooled) spatial feature maps. Non-local(\emph{NL}) \cite{wang2018non} takes pairwise relations/affinities as the weights to obtain an aggregated feature for refinement. \emph{SNL} is a simplified scheme of non-local \cite{cao2019gcnet}, which determines the weight for aggregation using only the source feature itself. NL ignores the mining of the global scope structural information from the relations and only uses them for weighted sum. In contrast, our RGA aims to mine from the relations. It is observed that the weights for aggregation in schemes \emph{NL} and \emph{SNL} are invariant to the target positions \cite{cao2019gcnet}. Thanks to the exploration of global structural information and its mining through learnable modeling function, our \emph{RGA-S} achieves the best performance, which is about 2\% better than the others in mAP accuracy on CUHK03(L).
%through a learnable modeling function

\begin{table}[t]
\centering
\caption{Performance (\%) comparisons of our attention and other approaches, applied on top of our baseline.}
\footnotesize
\resizebox{0.475\textwidth}{!}{
	\begin{tabular}{cccccc}
		\hline
		\multicolumn{2}{c}{\multirow{2}[4]{*}{Methods}} & \multicolumn{2}{c}{CUHK03 (L)} & \multicolumn{2}{c}{Market1501} \bigstrut\\
		\cline{3-6}    \multicolumn{2}{c}{} & R1 & mAP & R1 & mAP \bigstrut\\
		\hline
		Baseline & \multicolumn{1}{l}{ResNet-50} & 73.8 & 69.0 & 94.2 & 83.7 \bigstrut\\
		\hline
		\multirow{5}[2]{*}{Spatial}
		& \multicolumn{1}{l}{CBAM-S \cite{woo2018cbam}} & 77.3 & 72.8 & 94.8 & 85.6 \bigstrut[t]\\ 
		& \multicolumn{1}{l}{FC-S \cite{liu2018spatial}} & 77.0 & 73.0 & 95.2 & 86.2 \\
		& \multicolumn{1}{l}{NL \cite{wang2018non}} & 76.6 & 72.6 & 95.6 & 87.4 \\ 
		& \multicolumn{1}{l}{SNL \cite{cao2019gcnet}} & 77.4 & 72.4 & 95.7 & 87.3 \\
		& \multicolumn{1}{l}{RGA-S (Ours)} & \textbf{79.3} & \textbf{74.7} & \textbf{96.0} & \textbf{87.5} \bigstrut[b]\\
		\hline
		\multirow{4}[2]{*}{Channel} 
		& \multicolumn{1}{l}{SE \cite{hu2018squeeze}} & 76.3 & 71.9 & 95.2 & 86.0 \bigstrut[t]\\
		& \multicolumn{1}{l}{CBAM-C \cite{woo2018cbam}} & 76.9 & 72.7 & 95.3 & 86.3 \\
		& \multicolumn{1}{l}{FC-C \cite{liu2018spatial}} & 77.4 & 72.9 & 95.3 & 86.7 \\
		& \multicolumn{1}{l}{RGA-C (Ours)} & \textbf{79.3} & \textbf{75.6} & \textbf{95.9} & \textbf{87.9} \bigstrut[b]\\
		\hline
		\multirow{3}[2]{*}{Both} 
		& \multicolumn{1}{l}{CBAM-CS\cite{woo2018cbam}} & 78.0 & 73.0 & 95.0 & 85.6 \bigstrut[t]\\
		& \multicolumn{1}{l}{FC-S//C \cite{liu2018spatial}} & 78.4 & 73.2 & 94.8 & 85.0 \\
		& \multicolumn{1}{l}{RGA-SC (Ours)} & \textbf{81.1} & \textbf{77.4} & \textbf{96.1} & \textbf{88.4} \bigstrut[b]\\
		\hline
	\end{tabular}}%
	\label{tab:OtherAttention}%
	\vspace{-3mm}
\end{table}%

\begin{figure}[t]%[h]
	\begin{center}
		%\vspace{-3mm}
		\includegraphics[width=1\linewidth]{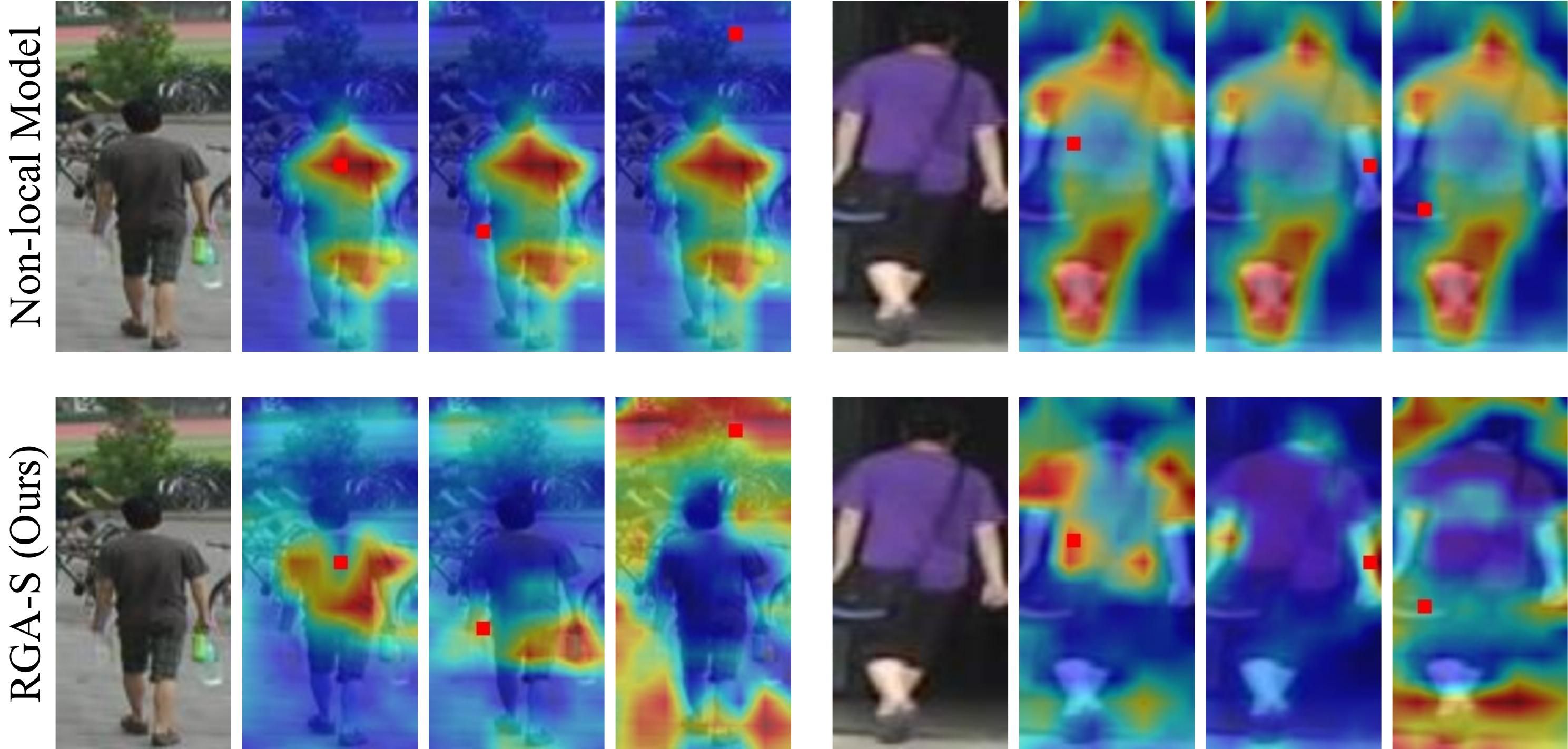}
	\end{center}
	\vspace{-3mm}
	\caption{Each three subimages visualize the connecting weights (relation values) from all positions w.r.t three target positions (marked by red squares), for non-local scheme (top row) and our RGA-S scheme (bottom row). For the color intensity, red indicates a large value while blue indicates a small one. We observe that the weights are invariant to the target positions for non-local model but adaptive in our RGA-S. For a target position, the positions with similar semantics usually have large relation values in our RGA-S, which reflects clustering-like patterns.}% in the semantics space
	%\vspace{-3mm}
	\label{fig:query_vis}
\end{figure}

To better understand the difference between the non-local \emph{NL} \cite{wang2018non} and our \emph{RGA-S}, we visualize their learned pairwise relation/affinity values with respect to three randomly selected target positions in Fig. \ref{fig:query_vis}. We find that the relation values are target position invariant for the non-local scheme (top row), which is similar to the observation made by Cao \etal \cite{cao2019gcnet}. In contrast, thanks to the learned modeling function applied on the relation vector and appearance feature, it drives the pairwise relation function (see Eq.~(1)) to better model the relations and makes the learned relations target position adaptive in our scheme (bottom row). For a target position, we observe the feature positions which have similar semantics are likely to have large relation/affinity values. This indicates that our attention model has mined helpful knowledge, \eg, clustering-like patterns in semantic space, from the relations for inferring attention.

\noindent\textbf{2) Channel attention.} In Squeeze-and-Excitation module (\emph{SE} \cite{hu2018squeeze}), they use spatially global average-pooled features to compute channel-wise attention, by using two fully connected (FC) layers with the non-linearity. In comparison with \emph{SE}, our \emph{RGA-C} achieves 3.0\% and 3.7\% gain in Rank-1 and mAP accuracy. \emph{CBAM-C} \cite{woo2018cbam} is similar to (\emph{SE}) \cite{hu2018squeeze} but it additionally uses global max-pooled features. Similarly, \emph{FC-C} \cite{liu2018spatial} uses a FC layer over spatially average pooled features. Before their pooling, the features are further embedded through $1\times1$ convolutions. Thanks to the exploration of pairwise relations, our scheme \emph{RGA-C} outperforms \emph{FC-C} \cite{liu2018spatial} and \emph{SE} \cite{hu2018squeeze}, which also use global information, by 1.9\% and 3.0\% in Rank-1 accuracy on CUHK03. On Market1501, even though the accuracy is already very high, our scheme still outperforms others.

\noindent\textbf{3) Spatial and channel attention.} When both spatial and channel attentions are utilized, our models consistently outperform using the channel attention alone or using the spatial attention alone.

\noindent\textbf{Parameters.} As shown in Table \ref{tab:params}, the number of parameters of the scheme \emph{RGA-S} is less than the \emph{NL} scheme, while the number of parameters of the scheme \emph{RGA-C} is about 2\% to 6\% larger than other schemes. 
% Table generated by Excel2LaTeX from sheet 'Sheet2'
%\vspace{-3mm}

\setlength{\tabcolsep}{3.5pt}
\begin{table}[t]%[htbp]
	\centering
	\caption{Number of parameters for different schemes (Million).}
	\footnotesize
	\resizebox{0.475\textwidth}{!}{
		\begin{tabular}{ccccccccc}
			\hline
			\multirow{2}[0]{*}{Baseline} & \multicolumn{4}{|c}{Spatial}   & \multicolumn{4}{|c}{Channel} \\
			& \multicolumn{1}{|c}{CBAM-S} & \multicolumn{1}{c}{FC-S}  & \multicolumn{1}{c}{NL} & \multicolumn{1}{c}{RGA-S} & \multicolumn{1}{|c}{CBAM-C} & \multicolumn{1}{c}{FC-C}  & \multicolumn{1}{c}{SE} & \multicolumn{1}{c}{RGA-C} \\
			\hline
			25.1 & \multicolumn{1}{|c}{26.1}  & 26.9 & 30.6 & 28.3 & \multicolumn{1}{|c}{26.4} & 26.4 & 27.6 & 28.1 \bigstrut[t]\\
			\hline
		\end{tabular}
	}%
	\label{tab:params}%
\end{table}%

% Table generated by Excel2LaTeX from sheet 'Sheet2'
% \setlength{\tabcolsep}{16.5 pt}
\begin{table}[t]%[htbp]
  \setlength{\tabcolsep}{11pt}
  \centering
  \caption{Influence of embedding functions on performance (\%).}
  \footnotesize
    %\resizebox{0.475\textwidth}{!}{
    \begin{tabular}{ccccc}
    % \toprule
    \hline
    \multirow{2}[0]{*}{Model} & \multicolumn{2}{c}{CUHK03 (L)} & \multicolumn{2}{c}{Market1501} \strut\\
    \cline{2-5}
          & R1    & mAP   & R1    & mAP \strut\\
    \hline
    \multicolumn{1}{c}{Baseline} & 73.8  & 69.0  & 94.2  & 83.7 \\
    \multicolumn{1}{c}{w/o Embedding} & 78.6  & 75.2  & 95.2  & 87.3 \\
    \multicolumn{1}{c}{Symmetric} & 79.4  & 75.2  & 95.6  & 87.4 \\
    \multicolumn{1}{c}{Asymmetric (Ours)} & 81.1  & 77.4  & 96.1  & 88.4 \\
    \hline
    % \bottomrule
    \end{tabular}%}%
  \label{tab:embedding}%
  \vspace{-2.5mm}
\end{table}%

\setlength{\tabcolsep}{9pt}
\begin{table*}[htbp]
\centering
\newcommand{\tabincell}[2]{\begin{tabular}{@{}#1@{}}#2\end{tabular}}
\caption{Performance (\%) comparisons with the state-of-the-arts on CUHK03, Market1501 and MSMT17.\protect\footnotemark[2]}
\footnotesize
%\resizebox{0.98\textwidth}{!}{
    \footnotesize
	\begin{tabular}{cccccccccc}
		\hline
		\multicolumn{2}{c}{\multirow{3}[3]{*}{Method}} & \multicolumn{4}{c}{CUHK03}    & \multicolumn{2}{c}{\multirow{2}[2]{*}{Market1501}} & \multicolumn{2}{c}{\multirow{2}[2]{*}{MSMT17}} \bigstrut[t]\\
		\multicolumn{2}{c}{} & \multicolumn{2}{c}{Labeled} & \multicolumn{2}{c}{Detected} & \multicolumn{2}{c}{} & \multicolumn{2}{c}{}  \bigstrut[t]\\
		\cline{3-10}
		\multicolumn{2}{c}{} & \multicolumn{1}{c}{Rank-1} & \multicolumn{1}{c}{\,mAP\,} & \multicolumn{1}{c}{Rank-1} & \multicolumn{1}{c}{\,mAP\,} & \multicolumn{1}{c}{Rank-1}  & \multicolumn{1}{c}{\,mAP\,}  & \multicolumn{1}{c}{Rank-1} & \multicolumn{1}{c}{\,mAP\,} \bigstrut[t]\\
		\hline
		% \multirow{2}[1]{*}{Basic ResNet-50} & \multicolumn{1}{l}{IDE~\cite{sun2017beyond}} & \multicolumn{1}{c}{43.8 } & \multicolumn{1}{c}{38.9 } & - & - & 85.3  & 68.5  & 73.2  & 52.8  & - & - \bigstrut[t]\\
		%       & \multicolumn{1}{l}{Gp-reid~\cite{almazan2018re}} & - & - & - & - & 92.2  & 81.2  & 85.2  & 72.8  & - & - \\
		% \hline
		\multirow{9}[1]{*}{\tabincell{l}{Attention \\ -based}} 
		& \multicolumn{1}{l}{MGCAM (CVPR18)~\cite{song2018mask}} & 50.1 & 50.2 & 46.7 & 46.9 & 83.8  & 74.3  & - & - \bigstrut[t]\\
		%& \multicolumn{1}{l}{\textcolor{blue}{MaskReID (ICME19)}~\cite{qi2018maskreid}} & - & - & - & - & 90.0  & 70.3  & - & - \\
		& \multicolumn{1}{l}{AACN (CVPR18)~\cite{xu2018attention}} & - & - & - & - & 85.9  & 66.9  & - & - \\
		& \multicolumn{1}{l}{SPReID (CVPR18)~\cite{kalayeh2018human}} & - & - & - & - & 92.5  & 81.3  & - & - \\
		& \multicolumn{1}{l}{HA-CNN (CVPR18)~\cite{li2018harmonious}} & 44.4 & 41.0 & 41.7 & 38.6 & 91.2  & 75.7  & - & - \\
		& \multicolumn{1}{l}{DuATM (CVPR18)~\cite{si2018dual}} & - & - & - & - & 91.4  & 76.6  & - & - \\
		& \multicolumn{1}{l}{Mancs (ECCV18)~\cite{wang2018mancs}} & 69.0 & 63.9 & 65.5 & 60.5 & 93.1  & 82.3  & - & - \\
		%& \multicolumn{1}{l}{MHN-6(IDE)~\cite{chen2019mixed}} & 69.7 & 65.1 & 67.0 & 61.2 & 93.6  & 83.6  & 87.5  & 75.2  & - & - \\
		& \multicolumn{1}{l}{MHN-6(PCB) (ICCV19)~\cite{chen2019mixed}} & 77.2 & 72.4 & 71.7 & 65.4 & 95.1  & 85.0  & - & - \\
		& \multicolumn{1}{l}{BAT-net (ICCV19)~\cite{fang2019bilinear}} & 78.6 & \underline{76.1} & 76.2 & \underline{73.2} & 95.1  & 84.7  & \underline{79.5} & \underline{56.8} \\
		%   & \multicolumn{1}{l}{MHN~\cite{chen2019mixed}} & \multicolumn{1}{c}{77.2} & \multicolumn{1}{c}{72.4} & \multicolumn{1}{c}{71.7} & \multicolumn{1}{c}{65.4} & 95.1  & 85 & \textbf{89.1} & 77.2  & - & - \\
		\hline
		\multirow{7}[1]{*}{Others}
		& \multicolumn{1}{l}{PCB+RPP (ECCV18)~\cite{sun2017beyond}} & 63.7  & 57.5  & - & - & 93.8  & 81.6  & \multicolumn{1}{c}{68.2 } & \multicolumn{1}{c}{40.4 } \bigstrut[t]\\
		& \multicolumn{1}{l}{HPM (AAAI19)~\cite{fu2018horizontal}} & 63.9  & 57.5  & - & - & 94.2  & 82.7  & - & - \\
		& \multicolumn{1}{l}{MGN(w flip) (MM19)~\cite{wang2018learning}} & 68.0  & 67.4  & 66.8  & 66.0  & \underline{95.7}  & 86.9  & - & - \\
		& \multicolumn{1}{l}{IANet (CVPR19)~\cite{hou2019interaction}} & - & - & - & - & 94.4  & 83.1  & 75.5 & 46.8 \\
		& \multicolumn{1}{l}{JDGL (CVPR19)~\cite{zheng2019joint}} & - & - & - & - & 94.8  & 86.0  & 77.2 & 52.3 \\
		& \multicolumn{1}{l}{DSA-reID (CVPR19)~\cite{zhang2019densely}} & \underline{78.9} & 75.2 & \underline{78.2} & 73.1 & \underline{95.7}  & \underline{87.6}  & - & - \\
		%& SAN       & \textbf{80.1} & \textbf{76.4} & \textbf{79.4} & \textbf{74.6}   & \textbf{96.1} & \textbf{88.0} & \underline{87.9}      & \underline{75.5}      & \textbf{79.2} & \textbf{55.7} \\
		& \multicolumn{1}{l}{OSNet (ICCV19)~\cite{zhou2019omni}} & - & - & 72.3  & 67.8  & 94.8  & 84.9  & 78.7 & 52.9 \\
		\hline
 		\multirow{2}[2]{*}{Ours } & \multicolumn{1}{l}{Baseline} & 73.8 & 69.0  & 70.5  & 65.5 & 94.2  & 83.7  & 75.7 & 51.5 \\
		& \multicolumn{1}{l}{\textbf{RGA-SC}} & \textbf{81.1} & \textbf{77.4} & \textbf{79.6} & \textbf{74.5} & \textbf{96.1} & \textbf{88.4} & \textbf{80.3} & \textbf{57.5} \\
		\hline
	\end{tabular}%}%
	\label{tab:SOA}%
	\vspace{-2mm}
\end{table*}%

\noindent\textbf{Influence of Embedding Functions.} We use asymmetric embedding functions (see Eq.~(\ref{eq:rij})) to model the directional relations ($r_{i,j}$, $r_{j,i}$) between node $i$ and node $j$. We compare it with symmetric embedding and no embedding in Table \ref{tab:embedding}. We observe that using the feature directly (w/o Embedding) or using symmetric embedding function also significantly outperforms \emph{Baseline} but is clearly inferior to using asymmetric embedding. It indicates that the main improvements come from our new design of relation-based attention learning, in which better relation modeling will deliver better performance. Using asymmetric embedding functions leaves more optimization space.

\noindent\textbf{Which ConvBlock to Add RGA-SC?} 
We compare the cases of adding the RGA-SC module to different residual blocks. The RGA-SC brings gain on each residual blocks and adding it to all blocks performs the best. Please refer to the supplementary for more details.

\begin{comment} %To save space 
% Table generated by Excel2LaTeX from sheet 'Where2Use'
\begin{table}[htbp]
\centering
\caption{Performance (\%) comparisons of adding RGA-CS modules to different stages of ResNet50 baseline.}
\footnotesize
\begin{tabular}{cccrc}
\hline
\multirow{2}[4]{*}{Model } & \multicolumn{2}{c}{CUHK03 (L)} & \multicolumn{2}{c}{Market1501} \bigstrut\\
\cline{2-5} & R1 & mAP & \multicolumn{1}{c}{R1} & mAP \bigstrut\\
\hline
\multicolumn{1}{l}{Baseline} & 73.8 & 69 & 94.2 & 83.7 \bigstrut\\
\hline
\multicolumn{1}{l}{ResBlock conv2\_x} & 76.5 & 71.9 & \multicolumn{1}{c}{95.3} & 86.1 \bigstrut[t]\\
\multicolumn{1}{l}{ResBlock conv3\_x} & 78.1 & 73.4 & \multicolumn{1}{c}{95.4} & 85.7 \\
\multicolumn{1}{l}{ResBlock conv4\_x} & 78.6 & 74.7 & \multicolumn{1}{c}{95.5} & 86.0 \\
\multicolumn{1}{l}{ResBlock conv5\_x} & 74.0 & 71.2 & \multicolumn{1}{c}{95.4} & 85.1 \bigstrut[b]\\
\hline
%\multicolumn{1}{l}{ResBlock conv2,3,4\_x} & 80.4 & 76.2 & \multicolumn{1}{c}{\textbf{95.8}} & 87.8 \bigstrut[t]\\
\multicolumn{1}{l}{ResBlock conv2,3,4,5\_x} & \textbf{80.4} & \textbf{76.5} & \multicolumn{1}{c}{\textbf{95.8}} & \textbf{88.1} \bigstrut[b]\\
\hline
\end{tabular}%
\label{tab:stages}%
\end{table}%
\end{comment}

\subsection{Comparison with the State-of-the-Art}

Table \ref{tab:SOA} shows the performance comparisons of our relation-aware global attention models (RGA-SC) with the state-of-the-art methods on three datasets. In comparison with the attention based approaches \cite{song2018mask,qi2018maskreid,xu2018attention,kalayeh2018human} which leverage human semantics (\eg foreground/background, human part segmentation) and those \cite{li2018harmonious,si2018dual,wang2018mancs} which learn attention from input images themselves, our \emph{RGA-SC} significantly outperforms them. On the three datasets CUHK03(L)/CUHK03(D), Market1501, and the large-scale MSMT17, in comparison with all other approaches, our scheme \emph{RGA-SC} achieves the best performance which outperforms the second best approaches by 1.3\%/1.3\%, 0.8\%, and 0.7\% in mAP accuracy, respectively. The introduction of our RGA-SC modules consistently brings significant gain over our 
\emph{Baseline}, \ie, \textbf{8.4}\%/\textbf{9.0}\%, \textbf{4.7}\%, and \textbf{6.0}\% in mAP accuracy, respectively.
%In comparison with all the approaches, our scheme \emph{RGA-SC} achieves the best performance on the three datasets CUHK03(L)/CUHK(D), Market1501, and the large-scale MSMT17, which outperforms the second best approaches by 1.3\%/1.3\%, 0.8\%, and 0.7\% in mAP accuracy, respectively.

%On DukeMTMC-reID, the approaches that have higher performance than ours are mostly trained on a higher input image resolution (i.e., 384$\times$128 or 336$\times$168). Experimentally, using the resolution of 384$\times$128 can improve our mAP by 3.0\%. Besides, \emph{MGN} \cite{wang2018learning} and \emph{HPM} \cite{fu2018horizontal} ensemble local features at multiple granularities, and \emph{MHN-6(PCB)} takes PCB \cite{sun2017beyond} as the baseline instead of ResNet-50 as ours. Compared to our baseline, we can achieve 3.7\% and 2.7\% improvements in mAP and Rank-1 respectively, which also demonstrates the effectiveness.
			
\subsection{Visualization of Attention}

\begin{figure}[t]
	\begin{center}
		%\vspace{-3mm}
		\includegraphics[width=1 \linewidth]{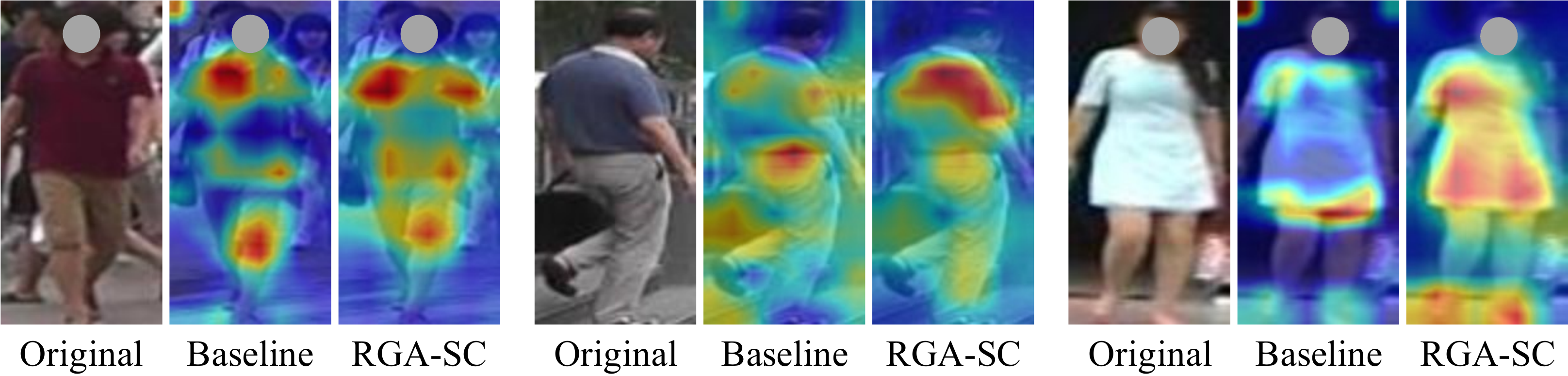}
	\end{center}
	\vspace{-3mm}
	\caption{Grad-CAM visualization according to gradient responses: \emph{Baseline} vs. \emph{RGA-SC}. }
	\vspace{-3mm}
	\label{fig:cam_vis}
\end{figure}
		
Similar to \cite{woo2018cbam}, we apply the Grad-CAM~\cite{selvaraju2017grad} tool to the baseline model and our model for the qualitative analysis. Grad-CAM tool can identify the regions that the network considers important. Fig.~\ref{fig:cam_vis} shows the comparisons. We can clearly see that the Grad-CAM masks of our RGA model cover the person regions better than the baseline model. The modulation function of our attention leads the network to focus on discriminative body parts.

We visualize the learned spatial attention mask in Fig.~\ref{fig:comparion}. The attention focuses on the person and ignores the background. In comparison with the attention approach of CBAM \cite{woo2018cbam} which does not exploit relations, our attention more clearly focuses on the body regions with discriminative information, which benefits from our mining of knowledge from the global scope structural information (where they present clustering-like patterns in semantic space (see the bottom row in Fig.~\ref{fig:query_vis}). Note that we observe that the head is usually ignored. That is because the face usually has low resolution and is not reliable for differentiating different persons. More visualization results including those on different layers can be found in the supplementary. 
%is not frontal or has
			
%------------------------------------------------------------------------
\section{Conclusion}

For person re-id, in order to learn more discriminative features, we propose a simple yet effective Relation-Aware Global Attention module which models the global scope structural information and based on this to infer attention through a learned model. The structural patterns provide some kind of global scope semantics which is helpful for inferring attention. Particularly, for each feature position, we stack the pairwise relations between this feature and all features together with the feature itself to infer the current position's attention. Such feature representation facilitates the use of shallow convolutional layers (\ie shared kernels on different positions) to globally infer the attention. We apply this module to the spatial and channel dimensions of CNN features and demonstrate its effectiveness in both cases. Extensive ablation studies validate the high efficiency of our designs and state-of-the-art performance is achieved.
%for person re-id.

\footnotetext[2]{We do not include results on DukeMTMC-reID \cite{zheng2017unlabeled} since this dataset is not publicly released anymore.}

%-------------------------------------------------------------------------
\section*{Acknowledgements}
This work was supported in part by NSFC under Grant U1908209, 61632001 and the National Key Research and Development Program of China 2018AAA0101400.

%-------------------------------------------------------------------------

{\small
\bibliographystyle{ieee_fullname}
\bibliography{main}

\begin{thebibliography}{10}\itemsep=-1pt

\bibitem{almazan2018re}
Jon Almazan, Bojana Gajic, Naila Murray, and Diane Larlus.
\newblock Re-id done right: towards good practices for person
  re-identification.
\newblock {\em arXiv preprint arXiv:1801.05339}, 2018.

\bibitem{barnes2009patchmatch}
Connelly Barnes, Eli Shechtman, Adam Finkelstein, and Dan~B Goldman.
\newblock Patchmatch: A randomized correspondence algorithm for structural
  image editing.
\newblock In {\em TOG}, volume~28, page~24. ACM, 2009.

\bibitem{buades2005non}
Antoni Buades, Bartomeu Coll, and J-M Morel.
\newblock A non-local algorithm for image denoising.
\newblock In {\em CVPR}, volume~2, pages 60--65, 2005.

\bibitem{Lefki2017non}
Antoni Buades, Bartomeu Coll, and J-M Morel.
\newblock Non-local color image denoising with convolutional neural networks.
\newblock In {\em CVPR}, 2017.

\bibitem{cao2019gcnet}
Yue Cao, Jiarui Xu, Stephen Lin, Fangyun Wei, and Han Hu.
\newblock Gcnet: Non-local networks meet squeeze-excitation networks and
  beyond.
\newblock {\em arXiv preprint arXiv:1904.11492}, 2019.

\bibitem{chen2019mixed}
Binghui Chen, Weihong Deng, and Jiani Hu.
\newblock Mixed high-order attention network for person re-identification.
\newblock In {\em ICCV}, pages 371--381, 2019.

\bibitem{chen2019relation}
Peihao Chen, Chuang Gan, Guangyao Shen, Wenbing Huang, Runhao Zeng, and Mingkui
  Tan.
\newblock Relation attention for temporal action localization.
\newblock {\em TMM}, 2019.

\bibitem{BM3DTIP07}
Kostadin Dabov, Alessandro Foi, Vladimir Katkovnik, and Karen Egiazarian.
\newblock Image denoising by sparse 3-d transform-domain collaborative
  filtering.
\newblock {\em TIP}, 16(8):2080--2095, 2007.

\bibitem{deng2009imagenet}
Jia Deng, Wei Dong, Richard Socher, Li-Jia Li, Kai Li, and Li Fei-Fei.
\newblock Imagenet: A large-scale hierarchical image database.
\newblock In {\em CVPR}, 2009.

\bibitem{efros1999texture}
Alexei~A Efros and Thomas~K Leung.
\newblock Texture synthesis by non-parametric sampling.
\newblock In {\em ICCV}, volume~2, pages 1033--1038. IEEE, 1999.

\bibitem{fang2019bilinear}
Pengfei Fang, Jieming Zhou, Soumava~Kumar Roy, Lars Petersson, and Mehrtash
  Harandi.
\newblock Bilinear attention networks for person retrieval.
\newblock In {\em ICCV}, pages 8030--8039, 2019.

\bibitem{fu2019sta}
Yang Fu, Xiaoyang Wang, Yunchao Wei, and Thomas Huang.
\newblock Sta: Spatial-temporal attention for large-scale video-based person
  re-identification.
\newblock AAAI, 2019.

\bibitem{fu2018horizontal}
Yang Fu, Yunchao Wei, Yuqian Zhou, Honghui Shi, Gao Huang, Xinchao Wang,
  Zhiqiang Yao, and Thomas Huang.
\newblock Horizontal pyramid matching for person re-identification.
\newblock {\em AAAI}, 2019.

\bibitem{glasner2009super}
Daniel Glasner, Shai Bagon, and Michal Irani.
\newblock Super-resolution from a single image.
\newblock In {\em ICCV}, pages 349--356. IEEE, 2009.

\bibitem{he2016deep}
Kaiming He, Xiangyu Zhang, Shaoqing Ren, and Jian Sun.
\newblock Deep residual learning for image recognition.
\newblock In {\em CVPR}, 2016.

\bibitem{hermans2017defense}
Alexander Hermans, Lucas Beyer, and Bastian Leibe.
\newblock In defense of the triplet loss for person re-identification.
\newblock {\em arXiv preprint arXiv:1703.07737}, 2017.

\bibitem{hou2019interaction}
Ruibing Hou, Bingpeng Ma, Hong Chang, Xinqian Gu, Shiguang Shan, and Xilin
  Chen.
\newblock Interaction-and-aggregation network for person re-identification.
\newblock In {\em CVPR}, pages 9317--9326, 2019.

\bibitem{hu2018squeeze}
Jie Hu, Li Shen, and Gang Sun.
\newblock Squeeze-and-excitation networks.
\newblock In {\em CVPR}, pages 7132--7141, 2018.

\bibitem{kalayeh2018human}
Mahdi~M Kalayeh, Emrah Basaran, Muhittin G{\"o}kmen, Mustafa~E Kamasak, and
  Mubarak Shah.
\newblock Human semantic parsing for person re-identification.
\newblock In {\em CVPR}, 2018.

\bibitem{li2018diversity}
Shuang Li, Slawomir Bak, Peter Carr, and Xiaogang Wang.
\newblock Diversity regularized spatiotemporal attention for video-based person
  re-identification.
\newblock In {\em CVPR}, pages 369--378, 2018.

\bibitem{li2014deepreid}
Wei Li, Rui Zhao, Tong Xiao, and Xiaogang Wang.
\newblock Deepreid: Deep filter pairing neural network for person
  re-identification.
\newblock In {\em CVPR}, 2014.

\bibitem{li2018harmonious}
Wei Li, Xiatian Zhu, and Shaogang Gong.
\newblock Harmonious attention network for person re-identification.
\newblock In {\em CVPR}, pages 2285--2294, 2018.

\bibitem{liu2018spatial}
Yiheng Liu, Zhenxun Yuan, Wengang Zhou, and Houqiang Li.
\newblock Spatial and temporal mutual promotion for video-based person
  re-identification.
\newblock In {\em AAAI}, 2019.

\bibitem{luo2016understanding}
Wenjie Luo, Yujia Li, Raquel Urtasun, and Richard Zemel.
\newblock Understanding the effective receptive field in deep convolutional
  neural networks.
\newblock In {\em NeurIPS}, pages 4898--4906, 2016.

\bibitem{qi2018maskreid}
Lei Qi, Jing Huo, Lei Wang, Yinghuan Shi, and Yang Gao.
\newblock Maskreid: A mask based deep ranking neural network for person
  re-identification.
\newblock {\em ICME}, 2019.

\bibitem{selvaraju2017grad}
Ramprasaath~R Selvaraju, Michael Cogswell, Abhishek Das, Ramakrishna Vedantam,
  Devi Parikh, and Dhruv Batra.
\newblock Grad-cam: Visual explanations from deep networks via gradient-based
  localization.
\newblock In {\em ICCV}, pages 618--626, 2017.

\bibitem{si2018dual}
Jianlou Si, Honggang Zhang, Chun-Guang Li, Jason Kuen, Xiangfei Kong, Alex~C
  Kot, and Gang Wang.
\newblock Dual attention matching network for context-aware feature sequence
  based person re-identification.
\newblock In {\em CVPR}, 2018.

\bibitem{song2018mask}
Chunfeng Song, Yan Huang, Wanli Ouyang, and Liang Wang.
\newblock Mask-guided contrastive attention model for person re-identification.
\newblock In {\em CVPR}, 2018.

\bibitem{suh2018part}
Yumin Suh, Jingdong Wang, Siyu Tang, Tao Mei, and Kyoung~Mu Lee.
\newblock Part-aligned bilinear representations for person re-identification.
\newblock In {\em ECCV}, 2018.

\bibitem{sun2017beyond}
Yifan Sun, Liang Zheng, Yi Yang, Qi Tian, and Shengjin Wang.
\newblock Beyond part models: Person retrieval with refined part pooling.
\newblock 2018.

\bibitem{szegedy2016rethinking}
Christian Szegedy, Vincent Vanhoucke, Sergey Ioffe, Jon Shlens, and Zbigniew
  Wojna.
\newblock Rethinking the inception architecture for computer vision.
\newblock In {\em CVPR}, 2016.

\bibitem{wang2018mancs}
Cheng Wang, Qian Zhang, Chang Huang, Wenyu Liu, and Xinggang Wang.
\newblock Mancs: A multi-task attentional network with curriculum sampling for
  person re-identification.
\newblock In {\em ECCV}, 2018.

\bibitem{wang2017residual}
Fei Wang, Mengqing Jiang, Chen Qian, Shuo Yang, Cheng Li, Honggang Zhang,
  Xiaogang Wang, and Xiaoou Tang.
\newblock Residual attention network for image classification.
\newblock In {\em CVPR}, pages 3156--3164, 2017.

\bibitem{wang2018learning}
Guanshuo Wang, Yufeng Yuan, Xiong Chen, Jiwei Li, and Xi Zhou.
\newblock Learning discriminative features with multiple granularities for
  person re-identification.
\newblock {\em ACM Multimedia}, 2018.

\bibitem{wang2018non}
Xiaolong Wang, Ross Girshick, Abhinav Gupta, and Kaiming He.
\newblock Non-local neural networks.
\newblock In {\em CVPR}, pages 7794--7803, 2018.

\bibitem{wang2018resource}
Yan Wang, Lequn Wang, Yurong You, Xu Zou, Vincent Chen, Serena Li, Gao Huang,
  Bharath Hariharan, and Kilian~Q Weinberger.
\newblock Resource aware person re-identification across multiple resolutions.
\newblock In {\em CVPR}, 2018.

\bibitem{wei2018person}
Longhui Wei, Shiliang Zhang, Wen Gao, and Qi Tian.
\newblock Person transfer {GAN} to bridge domain gap for person
  re-identification.
\newblock In {\em CVPR}, 2018.

\bibitem{woo2018cbam}
Sanghyun Woo, Jongchan Park, Joon-Young Lee, and In So~Kweon.
\newblock Cbam: Convolutional block attention module.
\newblock In {\em ECCV}, pages 3--19, 2018.

\bibitem{xu2018attention}
Jing Xu, Rui Zhao, Feng Zhu, Huaming Wang, and Wanli Ouyang.
\newblock Attention-aware compositional network for person re-identification.
\newblock In {\em CVPR}, pages 2119--2128, 2018.

\bibitem{yang2019attention}
Fan Yang, Ke Yan, Shijian Lu, Huizhu Jia, Xiaodong Xie, and Wen Gao.
\newblock Attention driven person re-identification.
\newblock {\em Pattern Recognition}, 86:143--155, 2019.

\bibitem{zhang2017alignedreid}
Xuan Zhang, Hao Luo, Xing Fan, Weilai Xiang, Yixiao Sun, Qiqi Xiao, Wei Jiang,
  Chi Zhang, and Jian Sun.
\newblock Alignedreid: Surpassing human-level performance in person
  re-identification.
\newblock {\em arXiv preprint arXiv:1711.08184}, 2017.

\bibitem{zhang2019nonlocal}
Yulun Zhang, Kunpeng Li, Kai Li, Bineng Zhong, and Yun Fu.
\newblock Residual non-local attention networks for image restoration.
\newblock In {\em ICLR}, 2019.

\bibitem{zhang2019densely}
Zhizheng Zhang, Cuiling Lan, Wenjun Zeng, and Zhibo Chen.
\newblock Densely semantically aligned person re-identification.
\newblock In {\em CVPR}, 2019.

\bibitem{zhao2017spindle}
Haiyu Zhao, Maoqing Tian, Shuyang Sun, Jing Shao, Junjie Yan, Shuai Yi,
  Xiaogang Wang, and Xiaoou Tang.
\newblock Spindle net: Person re-identification with human body region guided
  feature decomposition and fusion.
\newblock In {\em CVPR}, 2017.

\bibitem{zhao2017deeply}
Liming Zhao, Xi Li, Yueting Zhuang, and Jingdong Wang.
\newblock Deeply-learned part-aligned representations for person
  re-identification.
\newblock In {\em ICCV}, pages 3239--3248, 2017.

\bibitem{zheng2015scalable}
Liang Zheng, Liyue Shen, Lu Tian, Shengjin Wang, Jingdong Wang, and Qi Tian.
\newblock Scalable person re-identification: A benchmark.
\newblock In {\em ICCV}, 2015.

\bibitem{zheng2019joint}
Zhedong Zheng, Xiaodong Yang, Zhiding Yu, Liang Zheng, Yi Yang, and Jan Kautz.
\newblock Joint discriminative and generative learning for person
  re-identification.
\newblock In {\em CVPR}, pages 2138--2147, 2019.

\bibitem{zheng2017unlabeled}
Zhedong Zheng, Liang Zheng, and Yi Yang.
\newblock Unlabeled samples generated by gan improve the person
  re-identification baseline in vitro.
\newblock {\em arXiv preprint arXiv:1701.07717}, 2017.

\bibitem{zhong2017re}
Zhun Zhong, Liang Zheng, Donglin Cao, and Shaozi Li.
\newblock Re-ranking person re-identification with k-reciprocal encoding.
\newblock In {\em CVPR}, 2017.

\bibitem{zhong2017random}
Zhun Zhong, Liang Zheng, Guoliang Kang, Shaozi Li, and Yi Yang.
\newblock Random erasing data augmentation.
\newblock {\em arXiv preprint arXiv:1708.04896}, 2017.

\bibitem{zhou2019omni}
Kaiyang Zhou, Yongxin Yang, Andrea Cavallaro, and Tao Xiang.
\newblock Omni-scale feature learning for person re-identification.
\newblock {\em ICCV}, 2019.

\end{thebibliography}
}

\end{document}